# Med-BERT: pre-trained contextualized embeddings on large-scale structured electronic health records for disease prediction

Laila Rasmy*, M.S., Laila.Rasmy.GindyBekhet@uth.tmc.edu

Yang Xiang*†, Ph.D., Yang.Xiang@uth.tmc.edu

Ziqian Xie*, Ph.D., Ziqian.Xie@uth.tmc.edu

Cui Tao, Ph.D., Cui.Tao@uth.tmc.edu

Degui Zhi†, Ph.D., Degui.Zhi@uth.tmc.edu

School of Biomedical Informatics, University of Texas Health Science Center at Houston, U.S.

*Co-first authors with equal contributions

†Corresponding authors

**Abstract**

**Background:** Deep learning (DL)-based predictive models from electronic health records (EHRs) deliver good performance in many clinical tasks. Large training cohorts, however, are often required to achieve high accuracy, hindering the adoption of DL-based models in scenarios with limited training data size. Recently, bidirectional encoder representations from transformers (BERT) and related models have achieved tremendous successes in the natural language processing (NLP) domain. The pre-training of BERT on a very large training corpus generates contextualized embeddings that can be applied to smaller data sets with fine-tuning that can substantially boost their performance with these data sets. Because EHR data are analogous to text data, as both are sequential over a large vocabulary, we explore whether this "pre-training, fine tuning" paradigm can improve the performance of EHR-based predictive modeling.

**Objective:** To investigate whether pre-trained contextualized embeddings on large-scale structured EHRs can benefit downstream disease-prediction tasks and to share the pre-trained model and relevant tools with the public.

**Method:** We propose Med-BERT, which adapts the BERT framework for pre-training contextualized embedding models on structured EHR data. We improve the layer representations of BERT to make it more suitable for modeling the data structure of EHRs and design a domain-specific pre-training task to better capture the underlying semantics in the clinical data. Fine-tuning experiments are conducted on two disease-prediction tasks: (1) prediction of heart failure in patients with diabetes and (2) prediction of pancreatic cancer as well as on three cohorts from two EHR databases. The generalizability of the model on different sizes of fine-tuning training samples is tested. Further, the dependency relations among the EHRs of each patient as presented by different attention heads in the Med-BERT model are visualized.

**Results:** Med-BERT, pre-trained using a 28,490,650 patient EHR data set, substantially improves the prediction performance with the three fine-tuning cohorts, boosting the area under receiver operating characteristics (AUC) by 2.67–3.92%, 3.20–7.12%, and 2.02–4.71%, respectively, for three popular predictive models, i.e., GRU, Bi-GRU, and RETAIN. In particular, pre-trained Med-BERT considerably improved performance with very small fine-tuning training sets with only 300–500 samples, bringing their performance on par with a training set 10 times larger without pre-trained Med-BERT. Leveraging the parameters of Med-BERT, we also observe meaningful connections between clinical codes through dependency analysis.

**Conclusion:** Med-BERT is the first pre-trained contextualized embedding model that delivers a meaningful performance boost for real-world disease-prediction problems as compared to state-of-the-art models. Because it is especially helpful when only a small amount of data is available, it enables the pre-training, fine-tuning paradigm to be applied to solve various EHR-based problems. We share our pre-trained model to benefit disease-prediction tasks for researchers with small local training datasets. We believe that Med-BERT has great potential to help reduce data collection expenses and accelerate the pace of artificial intelligence (AI)-aided diagnosis.

# INTRODUCTION

Artificial intelligence (AI)-aided disease prediction has undergone considerable development in recent years [1-3]. At present, it can improve the precision of diagnosis, enable disease prevention by early warning, streamline clinical decision making, and reduce healthcare costs [4-7]. Powerful AI tools, advanced conventional machine learning [8-10], and deep-learning [11-14] approaches also have been widely applied in clinical predictive modeling and have gained numerous successes. Given enough training samples, deep-learning models can achieve comparable or even better performance than domain experts in the diagnosis of certain diseases [15-19]. One prerequisite of typical deep-learning-based methods is the availability of large and high-quality annotated datasets, which are used to cover the underlying complex semantics of the input domain as much as possible and to avoid the under-fitting of model training [20,21]. Big EHR data, however, often are not accessible for numerous reasons, including the limited number of cases for new or rare conditions; difficulty in data cleaning and annotation, especially if collected from different sources; and some governance issues that hinder the data acquisition [22].

Transfer learning was developed to address the issue whereby some representations were first pre-trained on large volumes of unannotated datasets and then further adapted to guide other tasks [23]. A recent trend in transfer learning is to use self-supervised learning over large general datasets to derive a general-purpose pre-trained model that captures the intrinsic structure of the data, which can be applied to a specific task with a specific dataset by fine-tuning. This pre-training-fine-tuning paradigm has been proven to be extremely effective in natural language processing (NLP) [24-30] and, recently, computer visions [31,32]. Among the models, bidirectional encoder representations from transformers (BERT) are one of the most popular ones for handling sequential inputs, e.g., texts, with numerous variations [33-39]. BERT also has been applied to the clinical domain, where there already were publicly available models pre-trained on clinical texts [33,34,40].

Structured EHRs, as a primary input source for disease prediction, offer rich and well-structured information that reflects the disease progression of each patient and are one of the most valuable resources for health data analysis [41,42]. Adapting the BERT framework to structured EHRs is a natural idea based on the analogy between natural language texts and EHRs; i.e., both texts and EHRs are sequential modalities for tokens from a large vocabulary. Generally, the tokens in language texts are "words," and the corresponding tokens in EHRs are medical codes (e.g., diagnosis code, medication code, procedure code). In addition, whereas the token orders in texts are natural and self-evident, the orders in EHRs are usually sequential specific to the clinical domain, such as the temporal relationship between visits.

In addition, adapting BERT to structured EHRs, however, is non-trivial due to the essential differences between EHRs and texts. For example, how to organize the EHRs to efficiently match the structured inputs of BERT is still uncertain, and whether there are any applicable domain-specific pre-training tasks is still under investigation. As a result, successful research in this field is still lacking.

To the best of our knowledge, there are only two relevant studies in literature: BEHRT [43] and G-BERT [44]. These models, however, are limited in the following ways. BEHRT aimed to develop pre-trained models to predict the existence of any medical codes in certain visits. It uses positional embeddings to distinguish different visits and adds an *age* layer to imply temporal orders. The authors' definition of the area under receiver operating characteristics (AUC), however, was a non-standard one, making it difficult to compare the findings with those of previous predictive modeling studies. G-BERT applied a graph neural network (GNN) model to expand the context of each clinical code through ontologies and jointly trained the GNN and BERT embeddings. G-BERT modified the masked language model (Masked LM) pre-training task into domain-specific ones, including maximizing the gap between the existing and non-existing codes and using different types of codes to predict each other. G-BERT's inputs, however, are all single-visit samples, which are insufficient to capture long-term contextual information in EHRs. In addition, the size for their pre-training dataset is not large, making it difficult to evaluate its full potential. Neither BEHRT nor G-BERT uses disease-prediction tasks as the evaluation of their pre-trained model by fine-tuning.

To alleviate the aforementioned issues and to evaluate a pre-trained contextualized embedding model specific to disease prediction, we designed Med-BERT, a variation of BERT for structured EHRs. We compare Med-BERT with BEHRT and G-BERT in Table 1. The main feature of Med-BERT is that it has a much larger vocabulary size and pre-training cohort size than do the other two models, which will help to provide a reality check of BERT-based models. In Med-BERT, larger cohort size and longer visit sequences will greatly benefit the model in learning more comprehensive contextual semantics. We also believe that, by using a large and publicly accessible vocabulary, i.e., International Classification of Diseases (ICD)-9 plus ICD-10, Med-BERT will likely be deployable to different institutions and clinical scenarios. Further, only Med-BERT tested the utility of the pre-trained model on an external data source (Truven), and the results showed that our model has good scalability.

Similar to BEHRT and G-BERT, we used code embeddings to represent each clinical code, visit embeddings to differentiate visits, and the transformer structure to capture the inter-correlations between codes. Within each visit, we defined *serialization embeddings* to denote the relative order of each code, whereas neither BEHRT nor G-BERT introduced code ordering within a visit. Specifically, we designed a domain-specific pre-training task *prediction of prolonged length of stay in hospital* (Prolonged LOS), which is a popular clinical problem that requires contextual information modeling to evaluate the severity of a patient's health condition according to the disease progression. We expect that the addition of this task can help the model to learn more clinical and more contextualized features for each visit sequence and facilitate certain tasks. Evaluations (fine-tuning)[1] were conducted on two disease-prediction tasks: *the prediction of heart failure among patients with diabetes* (DHF) and *the prediction of onset of pancreatic cancer* (PaCa), and three patient cohorts from two different EHR databases, Cerner Health Facts®[2] and Truven Health MarketScan®.[3] These tasks are different from the pre-training prediction tasks (Masked LM and Prolonged LOS) and, thus, are good tests for the generalizability of the pre-trained model. In addition, we chose these tasks because they capture more complexity, not merely the existence of certain diagnosis codes, and are based on phenotyping algorithms that further integrate multiple pieces of information, such as constraints on time window, code occurrence times, medications, and lab test values (see below).

Experiments were conducted in three steps: (1) test the performance gains by adding Med-BERT on three state-of-the-art predictive models; (2) compare Med-BERT with a pre-trained clinical

---

[1] We will use the more general term *evaluation* to describe the tasks and *fine-tuning* to describe the techniques throughout.
[2] https://sbmi.uth.edu/sbmi-data-service/data-set/cerner/
[3] https://sbmi.uth.edu/sbmi-data-service/data-set/ibm/

word2vec-style embedding [47]; and (3) test to see how much Med-BERT would contribute to disease predictions with different training sizes.

**Table 1**. Comparison of Med-BERT with BEHRT and G-BERT from multiple perspectives.

| Criteria | BEHRT | G-BERT | Med-BERT |
| --- | --- | --- | --- |
| Type of input code | Caliber code for diagnosis developed by a college in London | Selected ICD-9 code for diagnosis + ATC code for medication | ICD-9 + ICD-10 code for diagnosis |
| Vocabulary size | 301 | <4K | 82K |
| Pre-training data source | CPRD (primary care data) [45] | MIMIC III (ICU data) [46] | Cerner HealthFacts (general EHRs) |
| Input structure | Code + visit + age embeddings | Code embeddings from ontology + visit embeddings | Code + visit + code serialization embeddings |
| Pre-training sample unit | Patient's visit sequence | Single visit | Patient's visit sequence |
| Total number of pre-training patients | 1.6M | 20K | 20M |
| Average number of visits for each patient for pre-training | Not reported but > 5 | <2 | 8 |
| Pre-training task | Masked LM | Modified Masked LM | Masked LM + prediction of prolonged length of stay in hospital |
| Evaluation task | Diagnosis code prediction in different time windows | Medication code prediction | Disease predictions according to strict inclusion/exclusion criteria |
| Total number of patients in evaluation tasks | 699K, 391K, and 342K for different time windows | 7K | 50K, 20K, and 20K for three task cohorts |

Our primary contributions are summarized as follows:
1. This work is the first proof-of-concept demonstration that a BERT-style model for structured EHRs can deliver a meaningful performance boost in real-world-facing predictive modeling tasks.

2. We innovatively designed a domain-specific pre-training task that is prevalent among EHR data and effective in capturing contextual semantics.
3. This work is the first demonstration of significantly boosted performance over state-of-the-art methods on multiple clinical tasks with phenotyped cohorts.
4. This work is the first that presents generalizability by boosting the performance in a dataset (Truven) other than the training dataset (Cerner).
5. We observed promising results even when training on very few samples, demonstrating the enabling power of pre-trained models for clinical tasks for which limited training data are available.
6. We made the pre-trained models publicly available and provided a visualization tool to demonstrate the dependency semantics in EHRs, which we think will greatly facilitate researchers in both the informatics and clinical domains.

**METHODS**

*Data Preparation*

We extracted our cohorts from two databases. Cerner Health Facts® was used for both pre-training and evaluation tasks, and Truven Health MarketScan® was used only during evaluation. Appendix A the includes details of different cohort definitions and data sources. Finally, we had one cohort for pre-training from Cerner and three phenotyped cohorts for evaluation, two of which were from Cerner (DHF-Cerner and PaCa-Cerner) and one from Truven (PaCa-Truven). The descriptive analysis of the cohorts used is shown in Table 2.

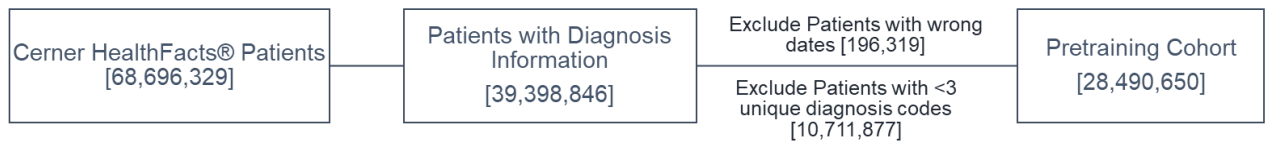

**Figure 1.** Selection pipeline for the pre-training cohort.

**Table 2**. Descriptive analysis of the cohorts.

| Characteristic | Pre-training | DHF-Cerner | PaCa-Cerner | PaCa-Truven |
| --- | --- | --- | --- | --- |
| Cohort size (*n*) | 28,490,650 | 672,647 | 29,405 | 42,721 |
| Average number of visits per patient | 8 | 17 | 7 | 19 |
| Average number of codes per patient | 15 | 33 | 14 | 18 |
| Vocabulary size | 82,603 | 26,427 | 13,071 | 7,002 |
| ICD-10 codes (%) | 33.8% | 13.3% | 20.7% | 0% |

*Med-BERT*

Figure 2 shows the Med-BERT structure with an input example (the dotted part). Three types of embeddings were taken as inputs. These embeddings were projected from diagnosis codes, the order of codes within each visit, and the position of each visit and named, respectively, *code embeddings*, *serialization embeddings*, and *visit embeddings*. Code embeddings are the low-dimensional representations of each diagnosis code; serialization embeddings denote the relative order of each code in each visit; and visit embeddings are used to distinguish each visit in the sequence.

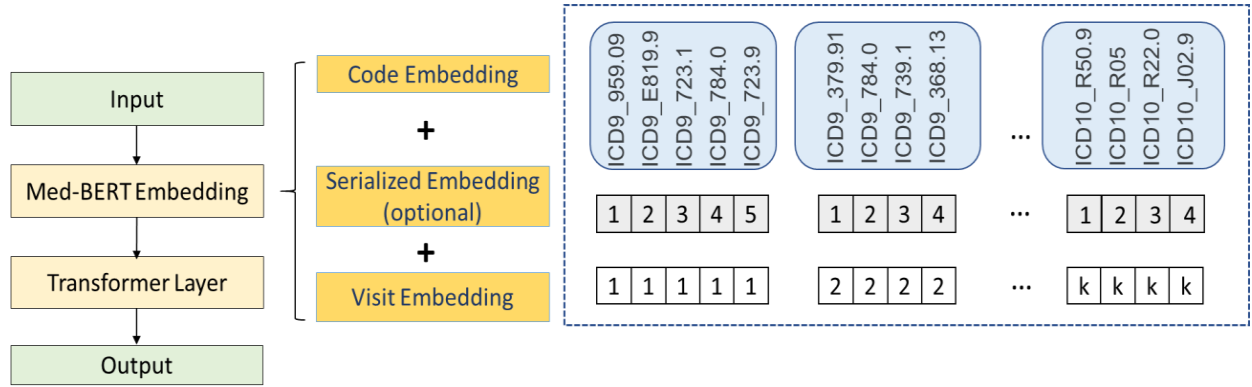

**Figure 2**. Med-BERT structure.

Unlike BERT, we did not use the specific tokens [*CLS*] and [*SEP*] at the input layer, attributed mainly to the differences in the input formats of EHRs and text. In BERT, there are only two adjacent sentences, and the token [*SEP*] behaves as a separator to distinguish the two sentences for the pre-training task of *next sentence prediction*. Next sentence prediction, however, was not involved in our tasks (as explained in the next subsection). We consider that the visit embeddings can separate well each visit and that adding [*SEP*] would only be redundant. In BERT, the token [*CLS*] was used mainly to summarize the information from the two sentences; however, EHR sequences are usually much longer; e.g., a sequence may contain 10 more visits, and simply using one token will inevitably lead to huge information loss. Therefore, for both the Prolonged LOS task and the downstream disease-prediction tasks where the information of the whole sequence is usually needed, we added a feed-forward layer (FFL) to average the outputs from all of the visits to represent a sequence, instead of using only a single token.

Similar to BERT, transformers also were employed to model the structures of diagnosis codes for each patient. Bidirectional transformer blocks in the transformer layer take advantage of multi-head attention operations to encode the codes and the visits [48].

*Pre-training*

Two tasks were involved in our pre-training step, on the code level and the patient level, respectively:

(1) *Masked Language Model* (*Masked LM*)

This model was directly inherited from the original BERT paper, which was used to predict the existence of code, given its context. In detail, there was an 80% chance that a code was replaced by [*MASK*], a 10% chance that the code was replaced by a random code, and another 10% chance that it was kept unchanged. This task is the core of the contextualized embedding model.

(2) *Prediction of prolonged length of stay (Prolonged LOS) in hospital*

For the classification task, instead of using the question-answer pairs, we decided to choose a clinical problem with a relatively high prevalence in our pre-training dataset and one that is not disease-specific to ensure better generalizability of our pre-trained model. The three most commonly used quality-of-care indicators, mortality, early readmission, and prolonged LOS, were selected and tested. Through testing that used different cohort definitions, we found that assessing each patient for whether a prolonged hospital visit (LOS >7 days) had ever occurred was associated with lower masking loss during the pre-training phase. Prolonged LOS involves the determination of whether a patient has experienced any prolonged LOS in his or her visit history based on the EHR sequence. A prolonged LOS is any hospital visit that is a longer than a seven-day stay, according to the difference between the admission and discharge dates. We used a simplified version of prolonged LOS prediction by targeting the patient level rather than the visit level to reduce the pre-training complexity.

The natural semantics of Prolonged LOS also can maximize the utility of the bidirectional structure of Med-BERT, as the task inherently indicates the severity of illness of each patient, which could be measured by the EHR from both directions (forward and backward), whereas tasks such as mortality always will be terminated at the last visit of the patient sequence, the input data of which can be constructed in only one direction. Similarly, early readmission will be based mainly on the pieces of visits involved, as they reflect the quality of care received and contribute more than the patient's health status in evaluating readmission.

*Evaluation*

The pre-trained model can be further fine-tuned to better fit downstream tasks, which is known as transfer learning. In our study, we conducted evaluations on two different disease-prediction tasks on three cohorts from two databases. The two tasks are *heart failure in diabetes patients* (*DHF-Cerner*) and *pancreatic cancer* (*PaCa-Cerner, PaCa-Truven*); the detailed cohort definitions are presented in the Supplementary Materials. Different from BEHRT and G-BERT, for which the evaluation tasks are simply the prediction of certain codes and are different from the tasks in pre-training, the definition of a certain disease is more complex, as it requires the phenotyping from multiple perspectives, e.g., the existence of certain diagnosis codes, drug prescriptions, procedures, lab-test values, and, sometimes, the frequency of events in predefined time windows. Therefore, we claim that our evaluation might be more realistic (compared with BEHRT) and more helpful in proving the generalizability of Med-BERT.

For each evaluation cohort, we randomly selected a subset of the original cohort and further split it into training, validation, and testing sets. The numbers of patients in the testing sets are all 5K for each task, and the sizes of training sets are 50K, 20K, and 20K for DHF-Cerner, PaCa-Cerner, and PaCa-Truven, respectively. The validation sets were set up according to a fixed ratio of the training sets, with 10% for the full training sets (e.g., 50K for DHF-Cerner) and 25% for other training sizes (e.g., training sets in Ex-3, introduced below). For performance measurement, we used AUC as our primary evaluation metric, which has been widely adopted by many previous studies of disease prediction [12,14,49].

For all three tasks, we conducted three experiments: (1) Ex-1: evaluate how Med-BERT can contribute to three state-of-the-art methods; (2) Ex-2: compare Med-BERT with one state-of-the-art clinical word2vec-style embedding, t-W2V (trained on the full Cerner cohort) [47]; and (3) Ex-3: investigate how much the pre-trained model can help in transfer learning, i.e., when only a proportion of training samples are available. For Ex-1 and Ex-2, we adopted GRU [50], Bi-GRU [51], and RETAIN [12] as our base models. Performances were compared through running the base models and adding pre-trained models as initializations and further fine-tuning, including t-W2V and Med-BERT. We also list the results by using Med-BERT only; i.e., only FFL was appended on top of the sequential output. For Ex-3, we selected different sizes of samples from the training data for each cohort for fine-tuning. Intuitively, the pre-trained model would be more helpful when the training size is smaller, as it helps inject a broader scope of knowledge. Therefore, the motivation of this stepwise experiment is to investigate how "small" our training set can be for achieving satisfying results with the addition of Med-BERT. For each training size (except the maximum size, which was randomly initialized 10 times), we conducted a random bootstrap sampling 10 times and reported the average AUC and standard deviation for each cohort.

*Implementation Details*

For pre-training we used mainly the BERT recommended parameters [29]. We had six attention heads, each with a 32-dimensional attention vector, and, thus, our hidden and embedding dimensions are 192. We used the default BERT optimizer, AdamWeight decay optimizer. We used the recommended learning rate of 5e-5, a dropout rate of 0.1, and 6 BERT layers. We set the maximum sequence length as 512 and masked only one diagnosis code per patient during Masked LM.

We used the Google TensorFlow code for pre-training (February 2019 version),[4] and, for fine-tuning, we converted the pre-trained model to the PyTorch version, using the Huggingface package (version 2.3) [52]. For pre-training, we used a single Nvidia Tesla V100 GPU of 32GB graphics memory capacity, and we trained the model for a week for more than 45 million steps, for which each step consists of 512 disease code (maximum sequence length) * 32 patients (batch size) . For the evaluation tasks, we used GPUs Nvidia GeForce RTX 2080 Ti of 12GB memory.

To facilitate reproducibility and benefit other EHR-based studies, we shared our pre-trained model as well as our visualization tool on https://github.com/ZhiGroup/Med-BERT. [The codes for all steps will be available to the public after the paper is accepted.]

**RESULTS**

*Performance boost of Med-BERT on fine-tuning tasks*

Table 3 presents the AUCs for Ex-1 on the three evaluation tasks. We tagged the top-three AUCs for each task as bold with different colors (**top-1**, top-2, top-3). For DHF-Cerner, it is notable that Bi-GRU+Med-BERT and RETAIN+Med-BERT obtain the best results and perform comparably, followed by Med-BERT_only and GRU+Med-BERT. For each base model, adding t-W2V (except GRU) will generally achieve better results, but adding Med-BERT improves the results much further. It is worth mentioning that, for those powerful deep-learning models, such as Bi-GRU and RETAIN, that can obtain over 80 on AUC with relatively large training data, e.g.,

---

[4] https://github.com/google-research/bert

50K samples, adding Med-BERT still makes considerable contributions, i.e., over 2.5%. Using Med-BERT_only can beat any model without Med-BERT.

**Table 3.** Average AUC values and standard deviations for the different methods for the three evaluation tasks.

| Model | DHF-Cerner | PaCa-Cerner | PaCa-Truven |
| --- | --- | --- | --- |
| *GRU* | 78.85 (1.65) | 78.05 (1.47) | 75.25 (0.47) |
| *GRU+t-W2V* | 74.72 (3.08) | 78.02 (0.89) | 74.57 (0.60) |
| *GRU+Med-BERT* | 82.77 (0.18) | 82.40 (0.16) | **79.27** (0.17) |
| *Bi-GRU* | 80.20 (0.21) | 77.76 (0.33) | 74.46 (0.19) |
| *Bi-GRU+t-W2V* | 80.72 (0.14) | 80.75 (0.16) | 75.23 (0.33) |
| *Bi-GRU+Med-BERT* | **83.15** (0.14) | 82.63 (0.22) | 79.17 (0.17) |
| *RETAIN* | 80.47 (0.22) | 78.83 (0.43) | 76.17 (0.25) |
| *RETAIN+t-W2V* | 82.38 (0.14) | 82.30 (0.18) | 78.60 (0.14) |
| *RETAIN+Med-BERT* | 83.14 (0.06) | 82.03 (0.17) | 78.19 (0.31) |
| *Med-BERT_only (FFL)* | 82.79 (0.12) | 82.41 (0.33) | 78.65 (0.2) |

For PaCa-Cerner, similar trends also were observed, whereby Bi-GRU+Med-BERT, Med-BERT_only, and GRU+Med-BERT outperformed methods without Med-BERT, and adding Med-BERT enhanced the AUCs of base models by 3–7%. For PaCa-Truven, the best AUC was obtained by GRU+Med-BERT, whereas the other Med-BERT-related models also have better results than do other models. On this Truven dataset, we still observed performance gains of 2.02–4.02%, although the average AUCs appear to be a bit lower than those on PaCa-Cerner. Nevertheless, we consider it enough to demonstrate that Med-BERT can be generalized well to a different dataset whose data distributions might be quite different from Cerner—the one it pre-trained on.

A clear trend reflected by the results of these three tasks is that the addition of Med-BERT was more helpful on simpler models, i.e., GRU > Bi-GRU > RETAIN, compared with the addition of t-W2V to the base models. The addition of t-W2V shows promising improvements on RETAIN and Bi-GRU, but neither helps nor hurts the performance of GRU. For RETAIN, the performance by adding t-W2V and Med-BERT is comparable on the PC tasks.

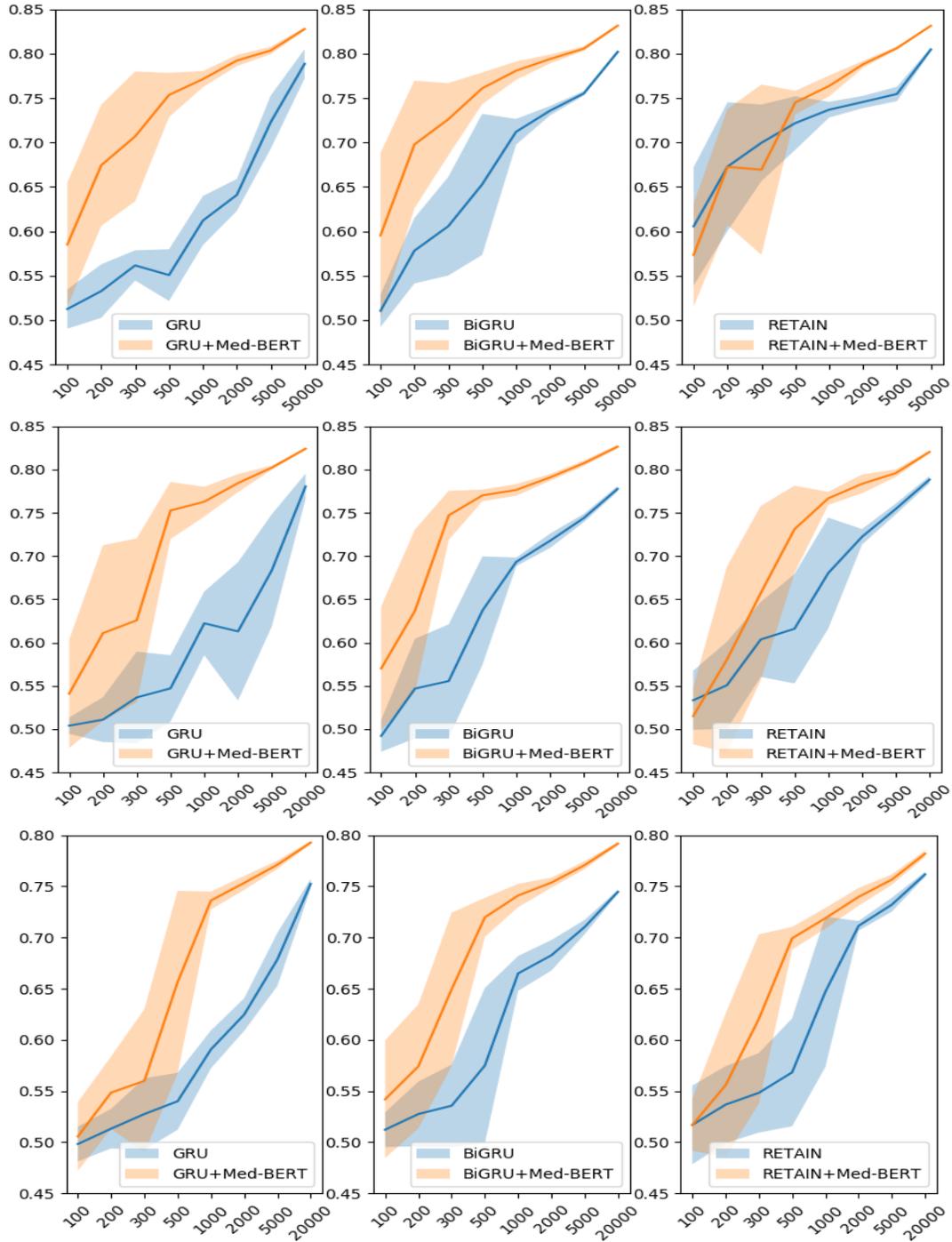

**Figure 3.** Prediction results for the evaluation sets by training on different sizes of data on DHF-Cerner (top), PaCa-Cerner (middle), and PaCa-Truven (bottom). The shadows indicate the standard deviations.

Figure 3 shows the results in the three cohorts on the transfer-learning paradigm. We used different training sizes to test how much Med-BERT can help boost the prediction performance of the base models by incorporating more contextual information as prior knowledge. In the line chart of DHF-Cerner, we notice that, without Med-BERT, it is difficult for GRU only to exceed 0.65 when given fewer than 2,000 training samples. The addition of Med-BERT, however, greatly increases the AUCs by about 20% and helps the model to reach 0.75, even when training on 500 samples. For Bi-GRU, considerable improvements also can be observed, but they are not as high as those for GRU. For RETAIN, Med-BERT seems to be more helpful when the training set contains more than 500 samples. For smaller training sizes, the standard deviations show heavy overlaps between the two curves, indicating that the results are not quite stable; thus, it is difficult to determine which model is absolutely better than the other.

For PaCa-Cerner, the results demonstrate large improvements by adding Med-BERT to GRU and Bi-GRU on almost all training sizes. In particular, for Bi-GRU, Med-BERT enables the AUC to reach 0.75 when training on only 300 samples. For RETAIN, when training on 100 samples, adding Med-BERT reduces the performance a bit, but the two curves overlap heavily based on standard deviations. The charts for PaCa-Truven show similar trends, but the overall AUC values are lower compared to those on PaCa-Cerner.

*Visualization of attention patterns in Med-BERT*

Med-BERT not only offers improvement for prediction accuracy but also enables prediction interpretation. It is interesting and meaningful to explore how the pre-trained model has learned using the complex structure and a huge volume of data. We show several examples of how codes are connected with each other according to the attention weights from the transformer layers, the core component of Med-BERT.

The bertviz tool [53] was adopted and improved to better visualize the attention patterns in each layer of the pre-trained model. We observed distinct patterns in different layers of the model. In the pre-trained model, among the six layers of the BERT transformer model, the connections of the first two layers are mostly syntactic, some attention heads are restricted within a visit, and some point to the same codes across different visits. In the middle two layers, some medically meaningful attention patterns that capture contextual and visit-dependent information emerge. For the final couple of layers, the attention patterns become diffused and difficult to interpret.

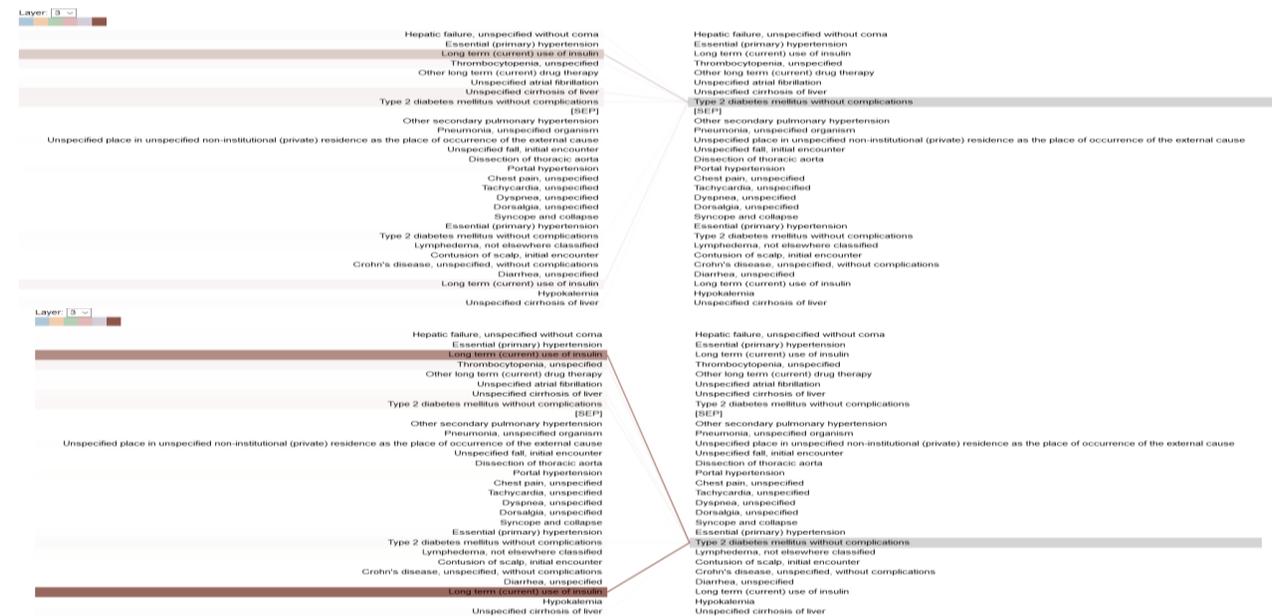

**Figure 4.** Example of different connections of the same code, "type 2 diabetes mellitus," in different visits.

Figure 4 is an example of the same code in different visits, showing different attention patterns. This demonstrates the ability of Med-BERT to learn contextualized representations. The earlier code for type 2 diabetes mellitus focuses mainly on the code for the long-term use of insulin within the same visit, but the later diabetes code focuses on the insulin code, both in the current and the previous visits. This could potentially indicate that the model learns the temporal relationship between visits through the segment embedding. More examples are provided in Appendix B.

The attention patterns of the fine-tuned model are different. The fine-tuned models express distinct task-dependent patterns across different layers, showing the generalizability and adaptability of the model for learning different levels of knowledge in the real-world scenarios. Figure 5 provides an example of the Med-BERT model fine-tuned on the DHF-Cerner dataset with attention converging onto several related codes in the second layer. Figure 6 is an example of the attention pattern in the fourth layer of the Med-BERT model fine-tuned on the PaCa-Cerner dataset, capturing the relevant correlation between diagnostic codes. Additional visualization patterns can be seen in the Supplementary Materials. We believe that these kinds of visualization patterns can help us to better understand the inner workings of the neural network model and to build trust and better communication of health information.

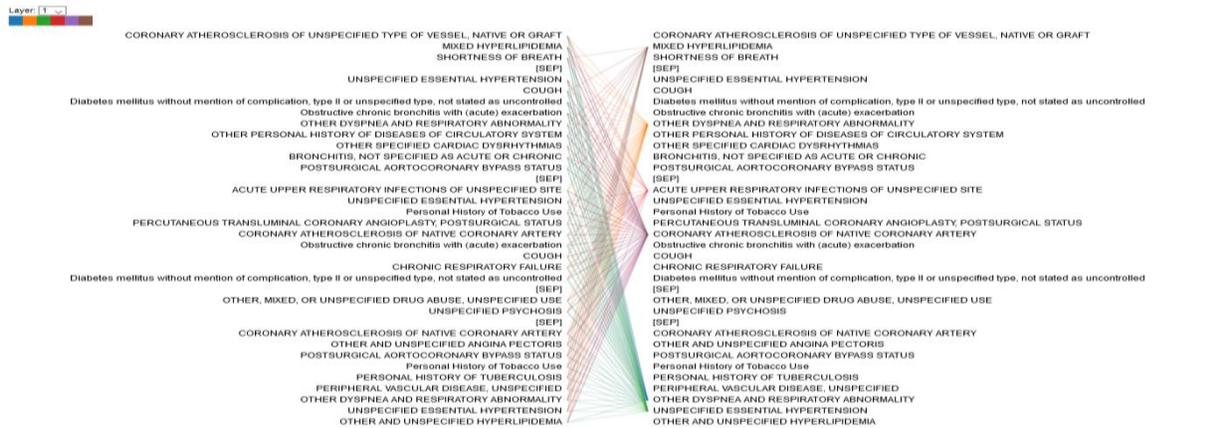

**Figure 5.** Example of the dependency connections in the DHF-Cerner cohort.

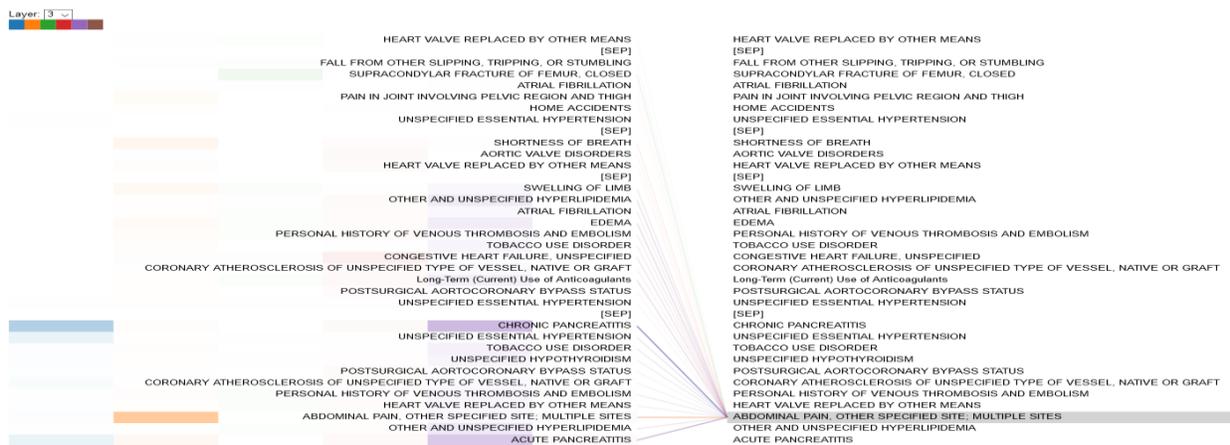

**Figure 6.** Example of the dependency connections in the PaCa-Cerner cohort.

**DISCUSSION**

Med-BERT shows its power in helping to improve the prediction performance on multiple tasks with different configurations, and it is particularly effective in the "extreme transfer learning" paradigms, i.e., fine-tuning on only several hundreds of samples. Deep-learning-based predictive models usually require at least thousands of samples. These models need to learn complex semantics through feeding samples that convey different underlying disease progressions and variational context information so that they can be capable of dealing with intricate unseen cases. However, most deep-learning algorithms are insufficient in modeling the data comprehensively due to their limitation in an in-depth understanding of the inputs. Pre-trained models can well address this issue by using more sophisticated structures to better capture the complex semantics of inputs, behaving as a knowledge container, and injecting the knowledge into new tasks. Similar to pre-trained models on other domains, Med-BERT, by using its bidirectional transformer and deep structures as well as big data, also have been shown in this study to be extremely helpful when transferring to new tasks.

Masked LM and Prolonged LOS were designed and included to reinforce the modeling of contextual information and to help collect sequential dependencies. Labels for both can be generated in an unsupervised way, i.e., without human annotations. In Masked LM, the goal is to predict a masked code using the sequential information from forward and backward directions. In Prolonged LOS, the goal is to determine whether a patient is associated with any visit that is a prolonged stay, which also relies on cumulative contexts. We believe that, by including the prediction tasks from both the code level and the patient (sequence) level, Med-BERT can further strengthen the representation learning of EHR sequences from different granularities.

Med-BERT also can improve several points when training on large cohorts. Intuitively, deep-learning methods can fit the data quite well if enough training data is available. We consider 50K and 20K as acceptable scales of samples for training satisfactory deep-learning models. When we added Med-BERT, however, significant improvements also could be observed. For example, RETAIN obtains the best performance on all the three tasks, but adding Med-BERT brings further improvements by 2.02–3.20%. In addition, for GRU and Bi-GRU, whose model structures are simpler than that of RETAIN, the improvements are much larger, which bring these simple models to a comparable level of RETAIN. Further, according to the results of Med-BERT_only, which also achieves good performance, we may conclude that Med-BERT will potentially release researchers from developing complex models for disease-prediction problems.

In comparison, t-W2V behaves more usefully in RETAIN but has limited contributions in Bi-GRU and GRU and even hurts the performance of GRU. A probable explanation is that t-W2V has limitations in modeling sequential information, considering its shallow structure and that it cannot be guaranteed to act well in all situations.

Ex-3 proves the effectiveness of transferring Med-BERT into realistic disease-prediction tasks. Most of the charts in Figure 3 reflect that Med-BERT has made an immense contribution to training base models on small samples. The only exception is RETAIN, where there are heavy overlaps of the two curves, e.g., under 500 samples in the third sub-chart of DHF-Cerner. A possible explanation is that RETAIN has a well-designed structure (two layers of self-attention) that is powerful in capturing some important features in these datasets and can get good results, whereas adding Med-BERT involves more parameters and more difficulties to fine-tuning. Note that the AUCs on those training sizes have large standard deviations, indicating that the data distribution might be uneven and that these results are not sufficiently stable.

Further in practice, Med-BERT will significantly help to reduce the data annotation cost, which can be seen in comparing the sizes of training samples required to achieve certain AUC levels. For example, in the first sub-chart of PaCa-Cerner in Figure 3, if we draw a horizontal line across the y-tick of 0.75, we will see a requirement of 500 samples for GRU+Med-BERT and over 5,000 samples for GRU only. Thus, Med-BERT brought the model performance on par with a training set almost 10 times larger. The data acquisition cost of these over 4,500 samples, which sometimes can be quite expensive, will be substantially saved by using Med-BERT. In this situation, with Med-BERT, researchers and clinicians are able to quickly get a general and acceptable understanding of the progressions of new diseases before collecting enough annotated samples.

The vocabulary of the current version of Med-BERT is the combination of ICD-9 and ICD-10 codes. Compared with BEHRT and G-BERT, our vocabulary is more acceptable and has broader coverage. We believe that it will greatly facilitate the mode transferability, as the ICD is the global health information standard recommended by the World Health Organization and is used by over 100 countries around the world.[5] This can be demonstrated in our PaCa-Truven evaluation, for which we tested models using a cohort extracted from a health insurance dataset.

There are still several limitations of the current work. First, we used only the diagnosis information in the ICD format at present. Second, we did not include the length of time intervals between visits in this study but, instead, used only the relative position, which may cause some temporal information loss. Third, we did not fully explore the order of concepts within each visit, and the current setting based on code priorities might not be sufficiently reliable. In the future, more research on designing different pre-training tasks will be conducted, and different types of evaluation tasks, other than disease prediction, also will be tested. We also plan to include other sources, such as time, medications, procedures, and laboratory tests, as inputs of Med-BERT. In addition, task-specific visualizations and interpretations are other areas that we plan to explore.

**CONCLUSION**

We proposed Med-BERT, a contextualized embedding model pre-trained on a large volume of structured EHRs, and further evaluated the model in disease-prediction tasks. Domain-specific input formats and pre-trained tasks were elaborately designed. Extensive experiments demonstrated that Med-BERT has the capacity to help boost the prediction performance of baseline deep-learning models on different sizes of training samples and can obtain promising results, even training on very few samples. The visualization module enabled us to look deeper into the underlying semantics of the data and working mechanisms of the model, in which we observed meaningful examples. Those examples were further verified by clinical experts, indicating that Med-BERT can well model the semantics among EHRs during both pre-training and evaluation. We believe that our model also can be beneficial in solving other clinical problems.

---

[5] https://www.who.int/classifications/icd/en/

**ACKNOWLEDGMENTS**

We are grateful for our collaborators, David Aguilar, MD, Masayuki Nigo, MD, and Bijun S. Kannadath, MBBS, MS, for the helpful discussions on cohorts definitions and results evaluation. This research was undertaken with the assistance of resources and services from the School of Biomedical Informatics Data Service, which is supported in part by CPRIT Grant RP170668. Specifically, we would like to acknowledge the use of Cerner HealthFacts® and the IBM Truven Marketplace™ datasets as well as the assistance provided by the UTHealth SBMI Data Service team to extract the data. The Nvidia GPU hardware is partly supported through Xiaoqian Jiang's UT star award. We are also grateful to the NVIDIA Corporation for supporting our research by donating a Tesla K40 GPU.

**Authors Contributions**

LR, YX, ZX, and DZ designed the methods. LR led the implementation of the methods, with substantial inputs from YX and ZX. YX and DZ led the design of experiments. LR conducted the experiments and produced results. ZX led the visualization. YX led the writing, with substantial inputs from LR, DZ, ZX, and CT. YX, DZ, and CT supervised the execution of the project. DZ initialized the conceptualization of the project.

**Funding Support**

CT and DZ are partly supported by the Cancer Prevention and Research Institute of Texas (CPRIT) Grant RP170668. LR is supported by UTHealth Innovation for Cancer Prevention Research Training Program Pre-Doctoral Fellowship (CPRIT Grant RP160015). CT and YX are supported by the National Institutes of Health (NIH) Grants R01AI130460 and R01LM011829.

**Disclaimer**

The content is solely the responsibility of the authors and does not necessarily represent the official views of the Cancer Prevention and Research Institute of Texas. The authors have no competing interests to declare.

# Supplementary Materials

## Appendix A: Data Extraction and Preparation

### A.1. Med-BERT pretraining cohort

Cerner Health Facts® (version 2017) is a de-identified EHR database that consists of over 600 hospitals and clinics in the United States, represents over 68 million unique patients, and includes longitudinal data from 2000 to 2017. The database consists of patient-level data, including demographics, encounters, diagnoses, procedures, lab results, medication orders, medication administration, vital signs, microbiology, surgical cases, other clinical observations, and health systems attributes. Data in Health Facts® is extracted directly from the EMR of hospitals with which Cerner has a data use agreement. Encounters may include pharmacy, clinical and microbiology laboratory, admission, and billing information from affiliated patient care locations. All admissions, medication orders and dispensing, laboratory orders, and specimens are date and time stamped, providing a temporal relationship between treatment patterns and clinical information. Cerner Corporation has established Health Insurance Portability and Accountability Act-compliant operating policies to establish de-identification for Health Facts®.

During the pretraining cohort data preprocessing, for each patient, we organized the visits in temporal order and ranked the diagnosis codes within each visit according to three criteria: (1) the diagnosis was flagged as *present* on admission; (2) the diagnosis was captured during the visit (e.g., hospitalization) or only at the billing phase; and (3) the diagnosis priority is provided by the Cerner database, indicating some priorities of the diagnoses, e.g., principal/secondary diagnosis (the priority is provided by the database, but it might not be a perfect priority ranking)

For each visit, we extracted the diagnosis codes (represented by ICD, Ninth Revision, Clinical Modification (ICD-9) and ICD, Tenth Revision, Clinical Modification (ICD-10)) and the length of stay in hospital. We then ranked the codes in each visit according to the above three criteria and determined the order by using (1) -> (2) -> (3) in sequence. We observed only very limited performance gains, however, by adding the code order during the evaluation, compared with randomly scattering the codes. Hence, we set it as a placeholder here and assume that more effective orders will be defined in the future.

Patients with fewer than three diagnosis codes in their records as well as those with wrong recorded time information, e.g., discharge date before admission date, were removed from the selection. In total, we had 28,490,650 unique patients (Figure 1), which were further separated into training, valid, and testing sets by the ratio of 7:1:2 on both the pre-training and evaluation phases.

### A.2. Diabetes heart failure cohort (DHF)

We originally identified 3,668,780 patients with at least one encounter with a diabetes diagnosis, based on the associated ICD-9/10 codes. We decided to exclude patients with any history of diabetes insipidus, gestational diabetes, secondary diabetes, neonatal diabetes mellitus, or type I diabetes mellitus (DM) from our cohort, as we focus on patients with type II DM and need to avoid any chance of wrong coding, taking into consideration that most of the EHR data are based on user manual entries and that there is a high associated chance of data entry mistakes. For the same reason, we decided to include patients who have more than one encounter with a diabetes diagnosis code. In addition, for type II DM patients, we verified that the patients' A1C reading is $\geq 6.5$ or that they are taking an antidiabetic agent, including metformin, chlorpropamide, glimepiride, glyburide, glipizide, tolbutamide, tolazamide, pioglitazone, rosiglitazone, sitagliptin, saxagliptin, alogliptin, linagliptin, repaglinide, nateglinide, miglitol, acarbose, or insulin.

For these cases, we identified patients with incidences of heart failure (using ICD-9 code equivalents, such as 428, or in 404.03, 404.13, 402.11, 404.11, 402.01, 404.01, 402.91, 398.91, 404.93, and 404.91, or ICD-10 code equivalents, such as I50%, or in I11.0, I09.81, I13.2, I97.13, I97.131, I13.0, and I97.130). In addition, we verified that the eligible cases are either prescribed a diuretic agent, had high B-type natriuretic peptide (BNP), or had been subjected to relevant procedures, including dialysis or an artificial heart-associated procedure. We included only those patients who reported heat failure (HF) at least 30 days after their first encounter with a type II DM code and excluded patients with only one HF encounter.

Further data cleaning included exclusion of patients with incorrect or incomplete data, for example, patients who were recorded as expired in between their first encounter and our event (first HF encounter for cases or last encounter for controls) as well as patients who are younger than 18 years old at their first diabetes diagnosis. The final cohort is shown in Supplementary Figure 1 and includes 39,727 cases and 632,920 controls.

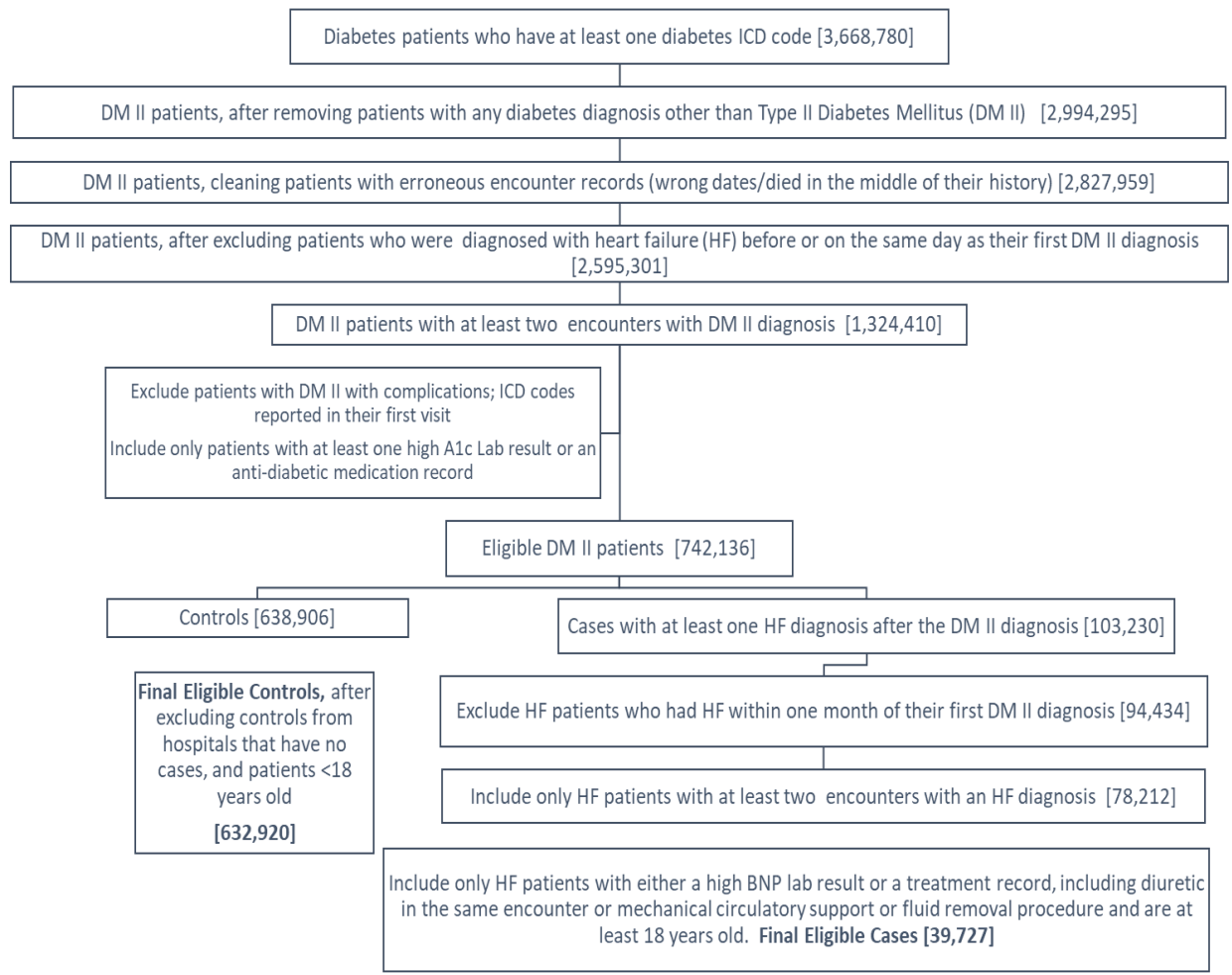

**Supplemental Figure 1.** Flowchart for the DHF cohort definition.

*A.3. Pancreatic cancer cohort (PaCa)*

Using ICD-9 codes that start with 157 and ICD-10 codes that start with C25, we originally identified around 45,000 pancreatic cancer patients from the Cerner HealthFacts dataset, of which 11,486 cases of individuals 45 years or older did not report any other cancer disease before their first pancreatic cancer diagnosis were eligible for inclusion in this cohort. Further details of the cohort definition are seen in Supplementary Figure 2.

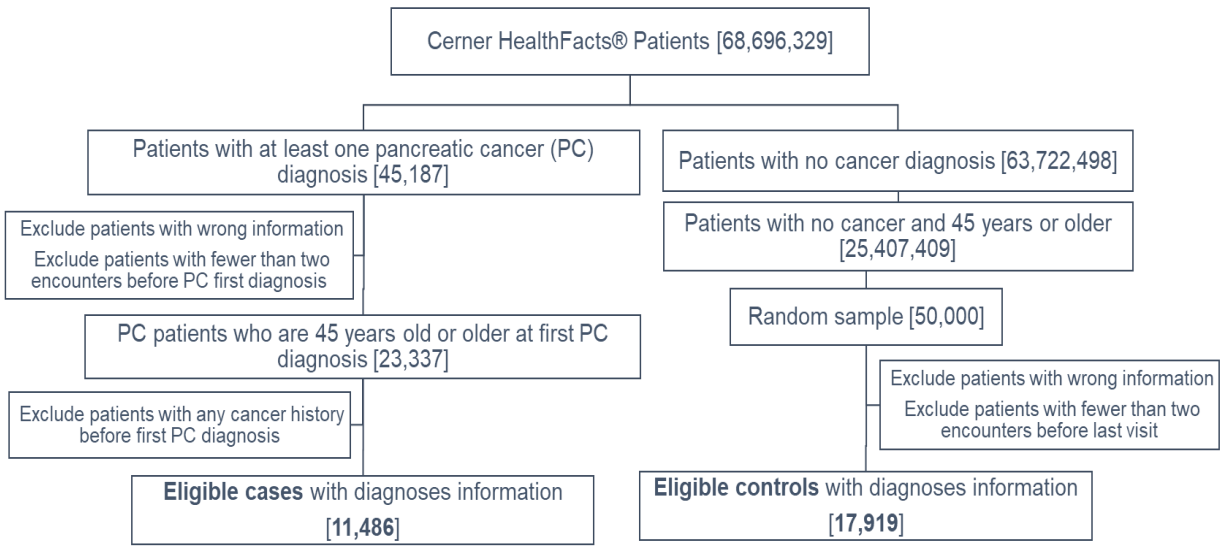

**Supplemental Figure 2.** Flowchart for the PC cohort definition.

Similarly, we extracted a cohort from Truven Health MarketScan® Research Databases for evaluation purposes. The Truven Health MarketScan® Research Databases (version 2015) are a family of research data sets that fully integrate de-identified patient-level health data (medical, drug, and dental), productivity (workplace absence, short- and long-term disability, and workers' compensation), laboratory results, health risk assessments, hospital discharges, and electronic medical records into datasets available for healthcare research. It captures person-specific clinical utilization, expenditures, and enrollment across inpatient, outpatient, prescription drug, and carve-out services. The annual medical databases include private-sector health data from approximately 350 payers. Historically, more than 20 billion service records are available in the MarketScan databases. These data represent the medical experience of insured employees and their dependents for active employees, early retirees, Consolidated Omnibus Budget Reconciliation Act (COBRA) continuees, and Medicare-eligible retirees with employer-provided Medicare Supplemental plans. Most of the diagnosis codes in Truven are ICD-9 codes, as the version of the database that we used is 2015, but the implementation of ICD-10 started in October 2015 [54].

**Appendix B: Additional Visualization Example**

Supplemental Figure 3 provides an example of the attention connections from the first three transformer layers. In the first layer, several heads show short-range attention patterns, and each token attends mainly to the nearby tokens that are within the same visit. In the second layer, some attention heads learn to make the correspondence between the same tokens. The third layer has the most interpretable patterns. A token in the third layer will focus strongly on other relevant tokens but mostly within the same visit. After the third layer, the attention becomes more diffuse and less explainable; however, there are still some heads that show long-range attention patterns.

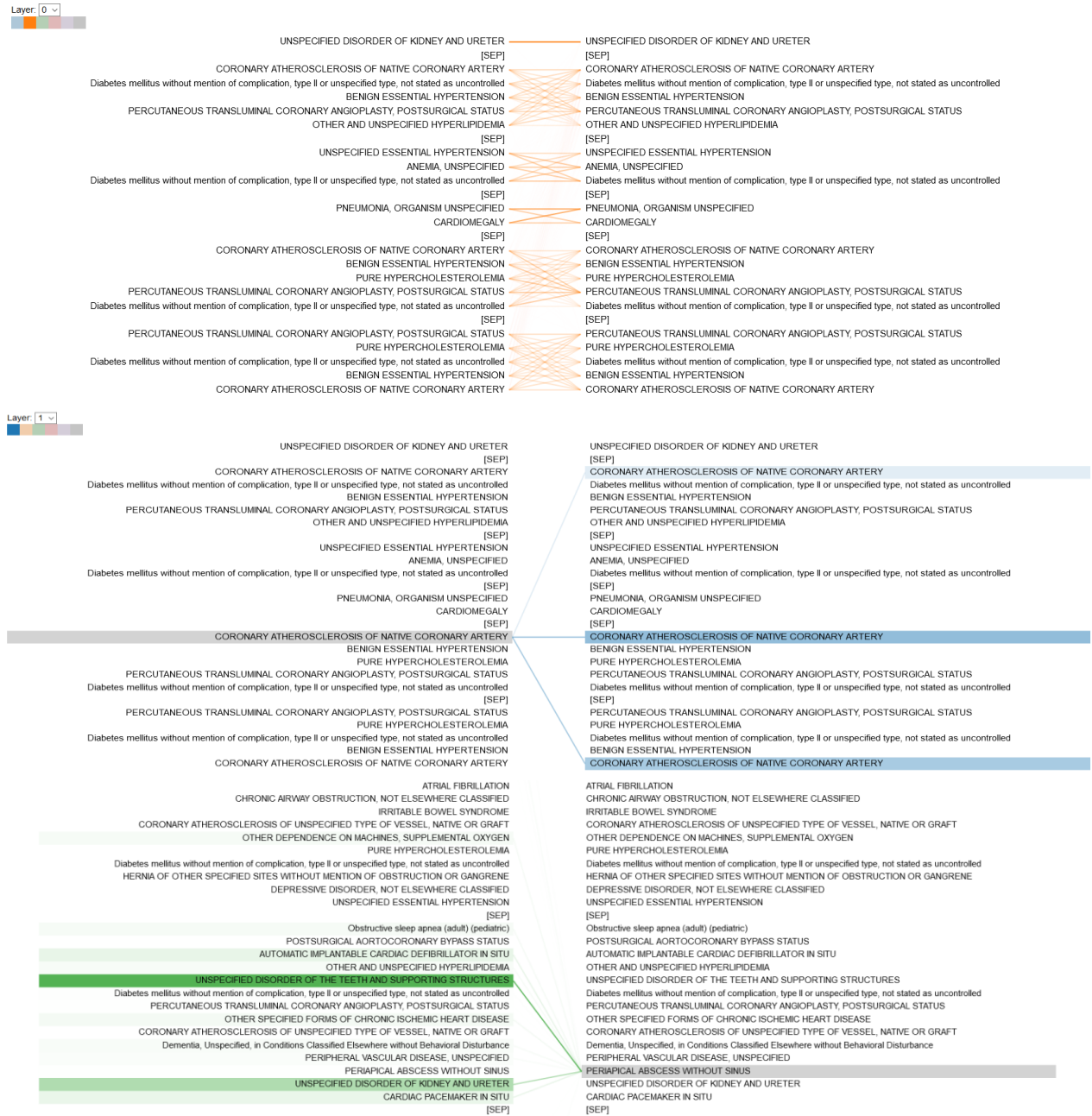

**Supplemental Figure 3.** Attention connections from the first three transformer layers (a top-down direction) of a sample patient sequence.